\documentclass[sigconf,anonymous=False]{acmart}
\usepackage{multirow}

\AtBeginDocument{%
  \providecommand\BibTeX{{%
    \normalfont B\kern-0.5em{\scshape i\kern-0.25em b}\kern-0.8em\TeX}}}

\setcopyright{acmcopyright}
\copyrightyear{2018}
\acmYear{2018}
\acmDOI{10.1145/1122445.1122456}

\acmConference[Woodstock '18]{Woodstock '18: ACM Symposium on Neural
  Gaze Detection}{June 03--05, 2018}{Woodstock, NY}
\acmBooktitle{Woodstock '18: ACM Symposium on Neural Gaze Detection,
  June 03--05, 2018, Woodstock, NY}
\acmPrice{15.00}
\acmISBN{978-1-4503-XXXX-X/18/06}

\begin{document}

\title{Hate Speech Detection in Clubhouse}

\author{Hadi Mansourifar}
\affiliation{%
  \institution{University of Houston}
  \streetaddress{4800 Calhoun Rd}
  \city{Houston}
  \state{Texas}
  \country{USA}
  \postcode{77004}
}
\email{hmansourifar@uh.edu}

\author{Dana Alsagheer}
\affiliation{%
  \institution{University of Houston}
  \streetaddress{4800 Calhoun Rd}
  \city{Houston}
  \state{Texas}
  \country{USA}
  \postcode{77004}
}
\email{dralsagh@central.uh.edu}

\author{Reza Fathi}
\affiliation{%
  \institution{University of Houston}
  \streetaddress{4800 Calhoun Rd}
  \city{Houston}
  \state{Texas}
  \country{USA}
  \postcode{77004}
}
\email{rfathi@uh.edu}
\author{Weidong Shi}
\affiliation{%
  \institution{University of Houston}
  \streetaddress{4800 Calhoun Rd}
  \city{Houston}
  \state{Texas}
  \country{USA}
  \postcode{77004}
}
\email{wshi3@central.uh.edu}
\author{Lan Ni}
\affiliation{%
  \institution{University of Houston}
  \streetaddress{4800 Calhoun Rd}
  \city{Houston}
  \state{Texas}
  \country{USA}
  \postcode{77004}
}
\email{lni2@central.uh.edu}
\author{Yan Huang}
\affiliation{%
  \institution{University of Houston}
  \streetaddress{4800 Calhoun Rd}
  \city{Houston}
  \state{Texas}
  \country{USA}
  \postcode{77004}
}
\email{yhuang63@central.uh.edu}
\settopmatter{printacmref=false}
\setcopyright{none}
\pagestyle{plain}
\renewcommand\footnotetextcopyrightpermission[1]{}
\begin{abstract}
  With the rise of voice chat rooms, a gigantic resource of data can be exposed to the research community for natural language processing tasks. Moderators in voice chat rooms actively monitor the discussions and remove the participants with offensive language. However, it makes the hate speech detection even more difficult since some participants try to find creative ways to articulate hate speech. This makes the hate speech detection challenging in new social media like Clubhouse. To the best of our knowledge all the hate speech datasets have been collected from text resources like Twitter. In this paper, we take the first step to collect a significant dataset from Clubhouse as the rising star in social media industry. We analyze the collected instances from statistical point of view using the Google Perspective Scores. Our experiments show that, the Perspective Scores can outperform Bag of Words and Word2Vec as high level text features.  
\end{abstract}



\maketitle

\section{Introduction}
"The explosive growth of Clubhouse [32], an audio-based social network buoyed by appearances from tech celebrities like Elon Musk and Mark Zuckerberg, has drawn scrutiny over how the app will handle problematic content, from hate speech to harassment and misinformation" [23]. However, in most of crowded rooms with presence of intellectual people, it's very hard to find overt act of aggressive speech since the moderators monitor and remove the impolite speakers. Consequently, hate speech is uttered in more articulated ways which is not compatible with traditional datasets already published in this domain. Beyond that, Audio presents a fundamentally different set of challenges for moderation than text-based communication. It’s more ephemeral and it’s harder to research and action,” said Discord’s chief legal officer, Clint Smith, in an interview [23].
Although the definition of hate speech [4,5,6,7,8,9] is still the same as any communication that disparages a person or a group on the basis of some characteristic such as race, color, ethnicity, gender, sexual orientation, nationality, religion or other characteristic [1,10,11,12,13], some people may convey it via more sophisticated ways. This issue is more frequently seen in voice-based social media in which people try to appear polite and educated. To the best of our knowledge, all the hate speech datasets have been collected from text resources like Twitter and audio resources are totally ignored so far. With the rise of voice chat rooms, a gigantic resource of data can be exposed to the research community for offensive language and hate speech detection. In this paper, we investigate the challenges of hate speech detection in Clubhouse. Following Israel-Palestine conflict in May 2021, an extensive discussions broke out about various aspects of this conflict from history and economy to religion and philosophy in number of Clubhouse rooms. Due to huge controversy surrounding this issue, the possibility of finding various instances of hate speech in such discussions was very likely. Recording more than 50 hours of discussions related to Israel-Palestine conflict in Clubhouse confirmed our initial assumption. Although the collected dataset is very imbalanced, positive instances include wide range of hate speech samples from highly toxic to seemingly polite ones. This diverse range of hate speech instances can provide the research community with a valuable resource for hate speech analysis in the future. Our collected data show that, a controversial topic can bring together many participants from across the globe in the Clubhouse. Although self-identification is very rare, we found 12 different nationalities among those who disclosed their identities. Also, we observed that such hot discussions in Clubhouse can encompass a wide range of themes from history to philosophy surrounding an international conflict. First, we extracted the text features using three different methods including Bag of Words [28], Word2Vec [29] and Perspective Scores [24]. Second, we used two different base classifiers including XGB [25] and LR [27] to compare the significance of extracted features. To extract the Perspective Scores we used Google Perspective API [24] and we investigated the significance of each score using ANOVA test for the task of binary classification. Our statistical analysis show that, some Perspective Scores including TOXICITY and IDENTITY-ATTACK have a significant impact on hate speech detection. Our contributions are as follows.
\vspace{-0.7mm}
\begin{itemize}
    \item We use voice-based social media to collect significant hate speech data.
    \item We publish a labelled dataset containing a diverse range of hate speech instances collected from the Clubhouse.
    \item Our experiments show that, Perspective Scores can play a crucial role in hate speech detection at voice-based social media.
\end{itemize}
The rest of paper is organized as follows. Section 2 reviews related works. Section 3 demonstrates data collection and annotation. Section 4 presents methodology. Section 5 provides the experimental results and finally section 6 concludes the paper.

\begin{figure*}[]
\centering
  \includegraphics[width=170mm]{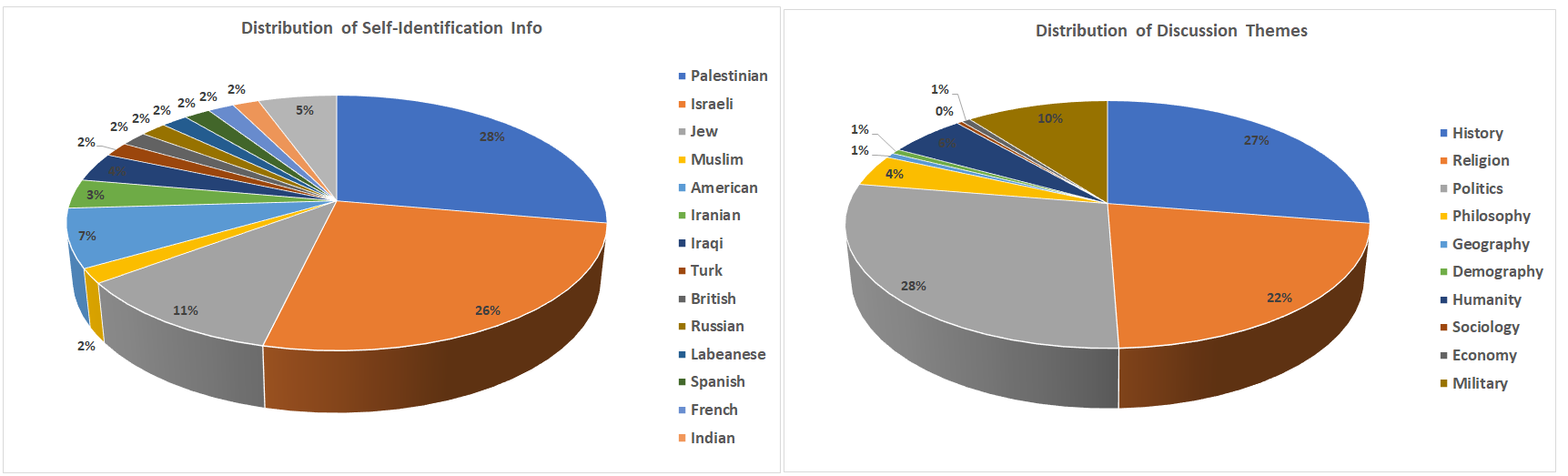}
  \caption{(A) Distribution of participants' identity according to self-identification info. (B) Distribution of discussions themes. }
  \label{ }
\end{figure*}

\begin{table*}[]
\caption{Hate speech samples collected from Clubhouse rooms discussing Israel-Palestine conflict.}
\begin{tabular}{|l|l|l|l|l|l|l|l|l|}
\hline
\textbf{Ex.} & \multicolumn{8}{c|}{\textbf{Hate Speech}} \\ \hline
1 & \multicolumn{8}{l|}{\textit{Do you support   Israel's right to exist? So, um that doesn't seem to me like a reasonable   question.}} \\ \hline
2 & \multicolumn{8}{l|}{\textit{I would just propose that Arabs sorry,   Palestinians who are citizens of Israel uh they are second class citizens.}} \\ \hline
3 & \multicolumn{8}{l|}{\textit{Israel is a nation for jews only, it's   not a nation for all people.}} \\ \hline
4 & \multicolumn{8}{l|}{\textit{I don't understand why the Israeli   people, we can't hear you, you're cutting off.}} \\ \hline
5 & \multicolumn{8}{l|}{\textit{when Palestinians are given freedom and   more rights, that suddenly it's going to be chaos and there's gonna be a lot   of violence.}} \\ \hline
6 & \multicolumn{8}{l|}{\textit{Well look how the Arabs treat each other,   how they treat their women as if kind of saying.}} \\ \hline
\end{tabular}
\end{table*}

\section{Related Work }
Our research contribution is to collect a new dataset from voice-based social media. To the best of our knowledge all the hate speech datasets [14,15,16] have been collected from text-based resources. In this section, we review some of the well known resources [17] and their characteristics.
\begin{itemize}
    \item DAVIDSON Dataset [12] : The dataset contains 24,802 tweets in English (5.77 \% labelled as hate speech, 77.43 \% as Offensive and 16.80 \%
as Neither) and was published in raw text format. They report collecting this data from Twitter using a lexicon from HateBase containing hateful words and phrases.

    \item WARNER Dataset [9]: The constituent data was collected from Yahoo News Group and URLs from the American Jewish Society. It contains
9000 paragraphs, manually annotated into seven (7) categories (anti-semitic, anti-black,
anti-Asian, anti-woman, anti-Muslim, anti-immigrant or other hate(anti-gay and anti-white)). The dataset is not publicly available.
    \item DJURIC Dataset [8]: This dataset was collected comments from the Yahoo Finance website. 56,280 comments were labeled as hateful while 895,456 labeled as clean from 209,776 users.
    \item QIAN Dataset [21]: This dataset was collected from Reddit and Gab including intervention responses written by humans. Their data
preserves the conversational thread as a way to provide context. From Reddit, they collected 5,020 conversations which includes a total of 22,324 comments labelled as hate or non-hate. 76.6 \% of the conversations contain hate speech while only 23.5 \% of the comments are labelled as hateful. They also collected 11,825 conversations containing 33,776 posts. 94.5 \% of the conversations contained hate speech while about 43.2 \% of the comments are labelled as hateful.
\item BENIKOVA Dataset [22] : contains 36 German tweets with 33 \% labelled as hate speech and 67 \% as non-hate speech.
\end{itemize}
\section{Data Collection and Annotation}
At the midst of Israel-Palestine conflict in May 2021, a wide range of discussions broke out from history to politics about various aspects of this conflict in number of Clubhouse rooms. Figure 1 shows the fraction of all sentences which directly pointing to one of these aspects including history, religion, politics, etc. Also, less than 1 \% of participants in the discussions self-identified themselves as Jew, Muslim, Palestinian, Israeli, Arab, etc. The distribution of identities is reported in part (a) of Figure 1. To collect the dataset about this topic, more than 50 hours of voice was recorded from 5 different rooms all discussing the Israel-Palestine conflict . Afterwards, all the voice data converted to the text format using Amazon Transcribe [30]. The raw text contains more than 11000 sentences. We labelled 468 instances including 122 hate speech instances and 346 normal instances. The annotation was done by two independent individuals and labels assigned with full agreement of both annotators.  
\begin{table*}[]
\centering
\caption{ANOVA test results on perspective scores of collected dataset. \\ Signif. codes:  0 ‘***’ 0.001 ‘**’ 0.01 ‘*’ 0.05 ‘.’ 0.1 ‘ ’ 1}
\begin{tabular}{lrrrrr}
  \hline
 & Df & Sum Sq & Mean Sq & F value & Pr($>$F) \\ 
  \hline
TOXICITY          & 1 & 56.66 & 56.66 & 1088.12 & 2e-16 *** \\ 
  SEVERE\_TOXICITY   & 1 & 0.39 & 0.39 & 7.41 & 0.0067 **\\ 
  IDENTITY\_ATTACK   & 1 & 7.77 & 7.77 & 149.13 & 2e-16 *** \\ 
  INSULT            & 1 & 0.54 & 0.54 & 10.35 & 0.0014 **\\ 
  PROFANITY         & 1 & 0.06 & 0.06 & 1.13 & 0.2880 \\ 
  THREAT            & 1 & 0.85 & 0.85 & 16.29 & 6.37e-05 ***\\ 
  SEXUALLY\_EXPLICIT & 1 & 0.00 & 0.00 & 0.00 & 0.9666 \\ 
  OBSCENE           & 1 & 0.00 & 0.00 & 0.04 & 0.8364 \\ 
  SPAM              & 1 & 0.06 & 0.06 & 1.23 & 0.2677 \\ 
  Residuals         & 457 & 23.80 & 0.05 &  &  \\ 
   \hline
\end{tabular}
\end{table*}

\subsection{What is in the Data?}
During the annotation, we understood that, some individuals articulate the hate speech in very sophisticated ways such that, the moderators could not find enough evidences to remove the participant from the room. Table 1 shows some of the hate speech instances collected from Clubhouse. We also observed that, the collected comments encompass a wide range of topics from history to economy as shown in part (b) of Figure 1. According to this chart, the most discussed topics are related to history, religion and politics. Surprisingly, we can find topics related to philosophy and economy in the comments which proves the capability of voice-based social media to bring together people from different backgrounds and point of views. We also observed that, such hot discussions over a controversial topic can attract people from different nationalities. Although less than 1 \% of participants disclosed their nationalities, limited self-identification info show a long list of nationalities as shown in part(a) of Figure 1.
\subsection{Public Dataset}
Our labelled dataset is publicly available in [31] but the raw text dataset which includes more than 11000 sentences is not publicly available. The raw text dataset is provided to valid academic requests submitted to the first author of this paper. Note that, the original voices would not be published to public and private requests.
\subsection{Challenges of collecting data from Clubhouse}
In this section, we report number of challenges of collecting data from Clubhouse.
\begin{itemize}
\item It's not that easy to find the target rooms discussing a specific topic. First, a number of active individuals must be followed to get access to a range of available rooms. Finding the target people and rooms is a time consuming process.
\item In crowded rooms most of the participants trying to be polite and educated. As a consequence, offensive language is very rare in the rooms with many participants. 
\item Presence of active moderators is a great hurdle for some individuals who wants to use offensive language.

\end{itemize}

\begin{table*}[]
\centering
\caption{ANOVA test results on interaction of collected dataset's Perspective Scores.}
\begin{tabular}{lrrrrr}
  \hline
 & Df & Sum Sq & Mean Sq & F value & Pr($>$F) \\ 
  \hline
TOXICITY                 & 1 & 56.66 & 56.66 & 1167.20 & 2e-16 *** \\ 
  SEVERE\_TOXICITY          & 1 & 0.39 & 0.39 & 7.95 & 0.0050  **\\ 
  IDENTITY\_ATTACK          & 1 & 7.77 & 7.77 & 159.97 & 2e-16 *** \\ 
  INSULT                   & 1 & 0.54 & 0.54 & 11.10 & 0.0009 ***\\ 
  PROFANITY                & 1 & 0.06 & 0.06 & 1.21 & 0.2711 \\ 
  THREAT                   & 1 & 0.85 & 0.85 & 17.47 & 3.50e-05 ***\\ 
  SEXUALLY\_EXPLICIT        & 1 & 0.00 & 0.00 & 0.00 & 0.9654 \\ 
  OBSCENE                  & 1 & 0.00 & 0.00 & 0.05 & 0.8306 \\ 
  SPAM                     & 1 & 0.06 & 0.06 & 1.32 & 0.2510 \\ 
  TOXICITY:IDENTITY\_ATTACK & 1 & 0.24 & 0.24 & 4.95 & 0.0266 *\\ 
  TOXICITY:SPAM            & 1 & 1.45 & 1.45 & 29.77 & 8.03e-08 ***\\ 
  TOXICITY:THREAT          & 1 & 0.06 & 0.06 & 1.16 & 0.2813 \\ 
  TOXICITY:INSULT          & 1 & 0.06 & 0.06 & 1.33 & 0.2492 \\ 
  Residuals                & 453 & 21.99 & 0.05 &  &  \\ 
   \hline
\end{tabular}

\end{table*}


\begin{table*}[]
\caption{Experimental results on collected dataset using 2 base classifiers and three feature extraction methods. Bold values represent the best results.}
\begin{tabular}{|c|c|c|c|c|c|}
\hline
\multirow{2}{*}{\textbf{Classifier}} & \multirow{2}{*}{\textbf{Features}} & \multicolumn{4}{c|}{\textbf{Performance Measure}} \\ \cline{3-6} 
 &  & Accuracy & Precision & Recall & F1 \\ \hline
\multirow{2}{*}{XGB} & \textit{Bag of Words} & 0.848 & 0.7905 & 0.5879 & 0.6634 \\ \cline{2-6} 
 & \textit{Word2Vec} & 0.8586 & 0.7991 & 0.5923 & 0.6684 \\ \cline{2-6} 
 & \textit{Perspective Scores} & \textbf{0.9528} & 0.9183 & \textbf{0.9052} & \textbf{0.9089} \\ \hline
\multirow{2}{*}{LR} & \textit{Bag of Words} & 0.8543 & 0.7918 & 0.61 & 0.6753 \\ \cline{2-6} 
 & \textit{Word2Vec} & 0.8629 & 0.7424 & 0.7233 & 0.7255 \\ \cline{2-6} 
 & \textit{Perspective Scores} & 0.9463 & \textbf{0.9452} & 0.8783 & 0.9057 \\ \hline
\end{tabular}
\end{table*}

\section{Methodology}
In this section, we present our approaches to analyse the collected data and train classifiers for hate speech detection. We used three different approaches as feature extraction including Bag of Words [28], Word2vec [18,29] and Perspective Scores [24]. Since Perspective Scores are less well-known comparing to traditional text features we demonstrate them in following section.
\subsection{Perspective Scores}
In order to extract high level features from hate speech instances, we use Perspective API [24]. Perspective API developed by Jigsaw and Google’s Counter Abuse Technology team as a part of the Conversation-AI project. The API provides several pre-trained models to compute several scores between 0 and 1 for different categories as follows [10].
\begin{itemize}

\item toxicity is a “rude, disrespectful, or unreasonable
comment that is likely to make people leave a discussion.”
\item severe toxicity is a “very hateful, aggressive, disrespectful comment or otherwise very likely to make a
user leave a discussion or give up on sharing their perspective.”
\item identity attack are “negative or hateful comments targeting someone because of their identity.”
\item insult is an “insulting, inflammatory, or negative comment towards a person or a group of people.”
\item profanity are “swear words, curse words, or other obscene or profane language”
\item threat “describes an intention to inflict pain, injury, or
violence against an individual or group.”
\end{itemize}
All the trained models use Convolutional Neural Networks (CNNs), trained with GloVe word embeddings [18] and fine-tuned during training on data from online sources such as Wikipedia and The New York Times [10].

\subsubsection{Analysis of Variance for Regression}
Analysis of Variance (ANOVA) consists of calculations that provide information about levels of variability within a regression model and form a basis for tests of significance [19]. We investigated the significance of Perspective Scores on hate speech detection task using ANOVA test. Our experiments show that, TOXICITY and IDENTITY-ATTACK scores have significant impact on hate speech detection as shown in Table 2. Furthermore, we examine the significance of features interaction. Table 3 shows that, there is a significant interaction between TOXICITY and IDENTITY-ATTACK.

\section{Experiments}
In this section, we review the detail settings of our experiments including base classifiers, feature extraction and performance measures.
\subsection{Feature Extraction}
We used three different feature extraction method as follows.
\begin{itemize}
\item Bag of words [26,28]: The ‘bag of words’ is the word vector for each instance in the dataset which is a simple word count for each instance where each position of the vector represents a word. We used ngram range of (1,2) which means that each vector is divided into single words and also pairs of consecutive words. We also set the max size of the word vector to 10,000 words.
\item Word2vec : Word2Vec [18,29] is one of the most popular technique to learn word embeddings using shallow neural network. It was developed by Tomas Mikolov in 2013 at Google. As the pre-trained model we used Google News corpus trained on 3 million words and phrases. this model provides 300-dimensional vectors as the transferred data.
\item Perspective Scores: They are high level features calculated by different trained classifiers. We passed all the records to Google Perspective API and collected 9 Perspective Scores per input vector as mentioned in previous section. We applied based classifiers without over-sampling [13,14,20] on the transferred vectors.
\end{itemize}
\subsection{Base Classifiers}
\begin{itemize}
\item XGB [25]: XGBoost is an implementation of gradient boosted decision trees designed for speed and performance. It has become popular in applied machine learning and Kaggle competitions in recent years.
\item LR [27]: It is commonly used for many binary classification tasks which uses logistic function and log odds to perform a binary classification task.
\end{itemize}

\subsection{Performance measures}
Classifier performance metrics are typically evaluated by a confusion matrix, as shown in following table. The rows are actual classes, and the columns are detected classes. TP (True Positive) is the number of correctly classified positive instances. FN (False Negative) is the number of incorrectly classified
positive instances. FP (False Positive) is the number of incorrectly classified negative instances. TN (True Negative) is the number of correctly classified negative instances. The three performance measures including precision, recall and F1 are defined by following formulae.\\
\begin{table}[h]
\begin{tabular}{|l|l|l|}
\hline
                         & \textbf{Detected Positive} & \textbf{Detected Negative} \\ \hline
\textbf{Actual Positive} & TP                         & FN                         \\ \hline
\textbf{Actual Negative} & FP                         & TN                         \\ \hline
\end{tabular}
\end{table}
\textbf{Recall} = TP/(TP+ FN) \\
\textbf{Precision} = TP/(TP+ FP) \\
\textbf{F1} = (2* Recall * Precision) /( Recall+ Precision)  \\

\subsection{Implementation}
To implement the feature extraction methods and base classifiers, we used python libraries including Sklearn, Pandas, etc. All the codes and three versions of labelled data corresponding to three types of extracted features including Bag of Word, Word2vec and Perspective Scores can be found in [31].
\subsection{Results}

In this section, we report the experimental results as shown in Table 4 where the best results are shown in bold per each performance measure . We tested two different base classifiers on features extracted from collected dataset using three different feature extraction methods. The experimental results can be summarized as follows.
\begin{itemize}
\item In both classifiers, the best results belong to Perspective Scores as the feature extraction method in terms of all performance measures.
\item XGB as base classifier and Perspective Score as feature extraction show the best results in terms of Accuracy, Recall and F1 score.
\item LR as base classifier and Perspective Score as feature extraction show the best results in terms of Precision.
\end{itemize}

\section{Conclusion}
In this paper, we introduced a significant dataset collected from an un-investigated media. We showed that, a voice-based social media like Clubhouse has incredible potential to expose very diverse range of data which embrace a wide range of themes from philosophy to military. It proves that, voice-based chat rooms would become a hub for collecting data for many researchers from different branches of natural language processing. According to self-identification info, participants from 12 different nationalities joined the discussions related to recent Israel-Palestine conflict.  We observed that, participants avoid offensive language to utter the hate speech and try to come up with more sophisticated comments to express hate speech. That's why the hate speech detection in Clubhouse is challenging. We tested three different feature extraction methods using two different base classifiers. Our experimental results showed that, Google Perspective Scores outperform the traditional feature extraction methods including Bag of Words and Word2Vec in terms of all tested performance measures. 


\end{document}